# Fully Dense UNet for 2D Sparse Photoacoustic Tomography Artifact Removal

Steven Guan, Amir A. Khan, Siddhartha Sikdar, and Parag V. Chitnis

*Abstract*— **Photoacoustic imaging is an emerging imaging modality that is based upon the photoacoustic effect. In photoacoustic tomography (PAT), the induced acoustic pressure waves are measured by an array of detectors and used to reconstruct an image of the initial pressure distribution. A common challenge faced in PAT is that the measured acoustic waves can only be sparsely sampled. Reconstructing sparsely sampled data using standard methods results in severe artifacts that obscure information within the image. We propose a modified convolutional neural network (CNN) architecture termed Fully Dense UNet (FD-UNet) for removing artifacts from 2D PAT images reconstructed from sparse data and compare the proposed CNN with the standard UNet in terms of reconstructed image quality.**

## I. Introduction

PHOTOACOUSTIC imaging (PAI) is an emerging hybrid technique for imaging optically-absorbing chromophores in a medium through the detection of acoustic waves generated via thermoelastic expansion [1]–[3]. It combines the high contrast of optical imaging with the resolution and penetration depth of ultrasound imaging and does not suffer from major drawbacks found in each technique alone [4]. In photoacoustic tomography (PAT), the acoustic pressure waves are measured using an array of detectors that enclose the sample and are commonly arranged in a spherical, cylindrical, or planar geometry [5]. The goal of PAT image reconstruction is to recover the initial pressure distribution from measurements along the detection boundary. This task is a well-studied inverse problem and can be solved using methods such as filtered back projection [6], Fourier methods [7], [8], model based [9], [10], and time reversal (TR) [11]–[13]. Among these methods, TR is considered to be the most robust and least restrictive because it works well for any arbitrary detection geometry and heterogenous media [14].

A common challenge faced in PAT is that the acoustic waves can only be sparsely sampled in the spatial dimension. Each discrete spatial measurement requires its own detector, and it may be infeasible to build an imaging system with a sufficiently large number of detectors due to practical and physical limitations [15]–[17]. Reconstructing sparsely sampled data using standard methods result in low quality images with severe artifacts. Iterative reconstruction methods can be used to reduce artifacts and improve image quality by incorporating prior knowledge such as smoothness, sparsity, and total variation constraints into the reconstruction process [17]–[20]. However, selecting appropriate constraints can be a challenging task, especially for images with complex spatial structures. Furthermore, iterative methods can be time consuming because they require repeated evaluations of the forward and adjoint operators.

Deep learning is an emerging research area, in which specialized artificial neural networks are used for pattern recognition and machine learning tasks [21]. In particular, convolutional neural networks (CNN) are widely used for imaging tasks such as classification and segmentation [22]–[25]. Recently, there has been an increasing interest in applying deep learning to sparse PAT reconstruction given its potential as a computationally efficient reconstruction method with comparable performance to state-of-the-art iterative methods [26]–[29].

Many applications of deep learning for sparse tomographic image reconstruction follows a post-processing approach, where an initial corrupted image is first reconstructed from the sensor data using a simple inversion step and then a CNN is applied as a post-processing step for removing artifacts and improving image quality. This approach has been successfully applied to CT, MRI, and PAT and shown to achieve comparable image quality to iterative methods [27], [30]–[32].

Another approach termed "model based learning and reconstruction" is to directly use the forward and adjoint operators in the reconstruction process with prior constraints learned from training data using a CNN [26]. Hauptmann et al applied this approach to PAT reconstruction and showed that it requires fewer iterations to converge and recovers higher quality reconstructions than iterative methods. Furthermore, it was demonstrated to outperform the post-processing approach but at the expense of additional computation time.

In this work, we follow the post-processing approach and propose a modified CNN architecture termed Fully Dense UNet

S. Guan is with the Bioengineering Department, George Mason University., Fairfax, VA 22031 USA. (e-mail: sguan2@gmu.edu) and The MITRE Corporation., McLean, VA, 22102 (e-mail: sguan@mitre.org). The author's affiliation with The MITRE Corporation is provided for identification purposes only and is not intended to convey or imply MITRE's concurrence with, or support for, the positions, opinions or viewpoints expressed by the author.

A. Khan, and P. Chitnis, S. Sikdar are with the Bioengineering Department, George Mason University, Fairfax, VA 22031 USA. (e-mail: akhan50@gmu.edu, pchitnis@gmu.edu, and ssikdar@gmu.edu).

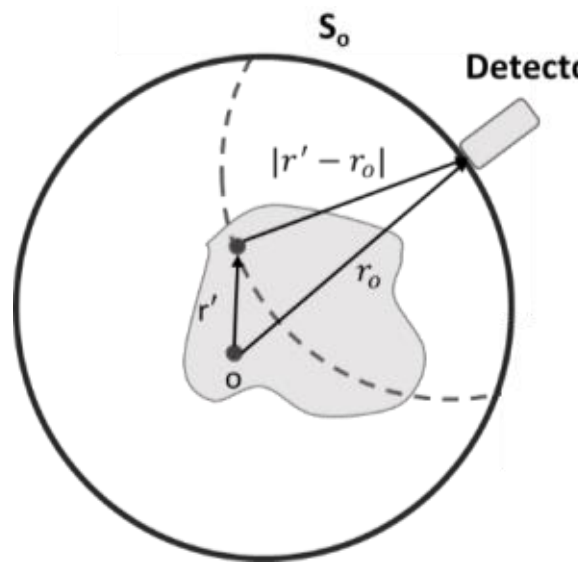

Fig. 1. Detector at position $r_o$ on the boundary $S_o$ measures the acoustic pressure emitted from a source located at $r'$. Adapted from [5].

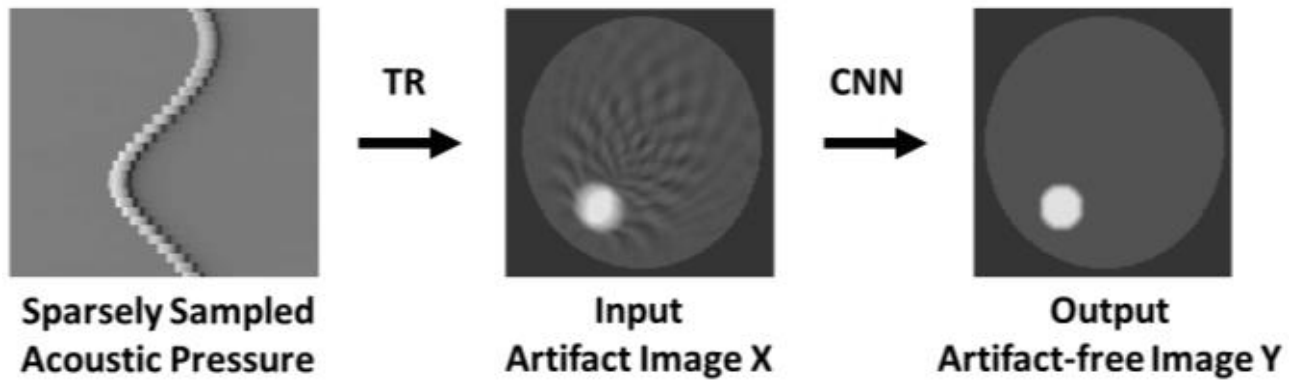

Fig. 2. Deep learning framework for 2D PAT image reconstruction. The sparsely sampled acoustic pressure is reconstructed into an image containing artifacts using time reversal. A CNN is applied to the artifact image $\boldsymbol{X}$ to obtain an approximately artifact free image $\boldsymbol{Y}$.

(FD-UNet) for removing artifacts in 2D PAT images reconstructed from sparse data. The FD-UNet incorporates dense connectivity into the contracting and expanding paths of the UNet CNN architecture. Dense connectivity mitigates learning redundant features and enhances information flow allowing for a more compact and superior CNN [23], [33], [34].

### A. Related Work

The UNet is the most widely used CNN architecture for applying deep learning with the post-processing approach in sparse tomographic image reconstruction [27], [29], [30]. It has many properties well-suited for artifact removal such as its use of multilevel decomposition and multichannel filtering [31]. Moreover, it has been demonstrated to perform comparatively well to iterative methods for sparse PAT image artifact removal on synthetic and experimental data [27], [29]. We build upon previous work and improve the post-processing approach by incorporating a recent advancement in CNN architecture design, namely dense connectivity, to achieve a CNN with superior performance. Compared to previous UNet implementations, we also apply batch normalization to accelerate the training process [35], [36].

The UNet with dense connectivity termed "DD-Net" has been previously used for sparse-view CT reconstruction and was shown to outperform iterative methods [37]. While the FD-UNet also uses dense connectivity, there are several differences in implementation. 1) The DD-Net includes dense connectivity only in the contracting path of the UNet. Whereas, the FD-UNet includes dense connectivity in both the contracting and expanding paths. This strategy enables the benefits of dense connectivity to be leveraged throughout the entire network. 2) In the DD-Net, the dense block "growth rate" hyperparameter remains constant throughout the network. In the FD-UNet, this hyperparameter is updated throughout the CNN to improve computational efficiency. To the best of our knowledge, this is the first work applying the UNet with dense connectivity for removing artifacts in sparse PAT image reconstruction.

## II. Methods

In PAT, the sample is irradiated by a short laser pulse $\delta(t)$ which leads to the generation of an initial acoustic pressure via thermo-elastic expansion [1]. For effective PAT signal generation, the laser pulse duration is typically only several nanoseconds in order to satisfy the thermal and stress confinement thresholds [3]. Given that these constraints are met, thermal diffusion and volume expansion during laser illumination is negligible, and the initial acoustic pressure can be written as

$$p_o(r) = \Gamma(r)A(r) \qquad (1)$$

where $A(r)$ is the spatial absorption function and $\Gamma(r)$ is the Grüneisen coefficient describing the conversion efficiency from heat energy to pressure [2]. The acoustic pressure wave $p(r,t)$ at position $r$ and time $t$ satisfies the following wave equation, in which $c$ is the speed of sound [6].

$$\left(\nabla^2 - \frac{1}{c^2}\frac{\partial^2}{\partial t^2}\right)p(r,t) = -p_o(r)\frac{d\delta(t)}{dt} \qquad (2)$$

Acoustic detectors at position $\boldsymbol{r_o}$ are located on a measurement surface $S_o$ that encloses the sample as seen in Fig. 1. Each detector along the surface measures a time-dependent signal of the emitted pressure wave over a period of time $T$. If a sufficiently large number of detectors are used then standard reconstruction techniques would yield an essentially artifact-free image [15]. In the 2D case, an image $\boldsymbol{X} \in \mathbb{R}^{d\times d}$ would require $M \geq \pi d$ detectors to satisfy Shannon's sampling theory. However, in most practical applications there are $M \ll \pi d$ detectors leading to a reconstructed image containing artifacts.

### A. Deep Learning Framework

As shown in Fig. 2., the sparsely sampled acoustic pressure is initially reconstructed using TR into an image $\boldsymbol{X}$ containing artifacts. The CNN is then applied to correct the undersampling artifacts in image $\boldsymbol{X}$ to obtain an approximately artifact-free image $\boldsymbol{Y}$. This task can be formulated as a supervised learning problem, in which the goal is to learn a restoration function that maps an input image $\boldsymbol{X}$ to the desired output image $\boldsymbol{Y}$ [27]. Other reconstruction methods can be used in place of TR to reconstruct the initial artifact image $\boldsymbol{X}$ from sensor data. TR was chosen for this work because it can be easily adapted for any sensor configuration, provides a good initial reconstruction, and is computationally inexpensive relative to iterative methods.

### B. Proposed FD-UNet Architecture

As seen in Fig. 3., the input image $\boldsymbol{X}$ undergoes a multilevel decomposition in the contracting path of the FD-UNet, where the spatial dimensions of the feature maps are repeatedly reduced via a max-pooling operator [22], [31], [38]. This strategy enables the CNN to efficiently learn local and global features relevant for artifact removal at various spatial scales [39]. In the following expanding path, the learned feature-maps are spatially upsampled via a deconvolution operator and combined to produce an output image $\boldsymbol{Y}$ with the same dimensions as the input image $\boldsymbol{X}$. Deconvolution can be thought as the reverse of convolution and is essentially a transposed convolution.

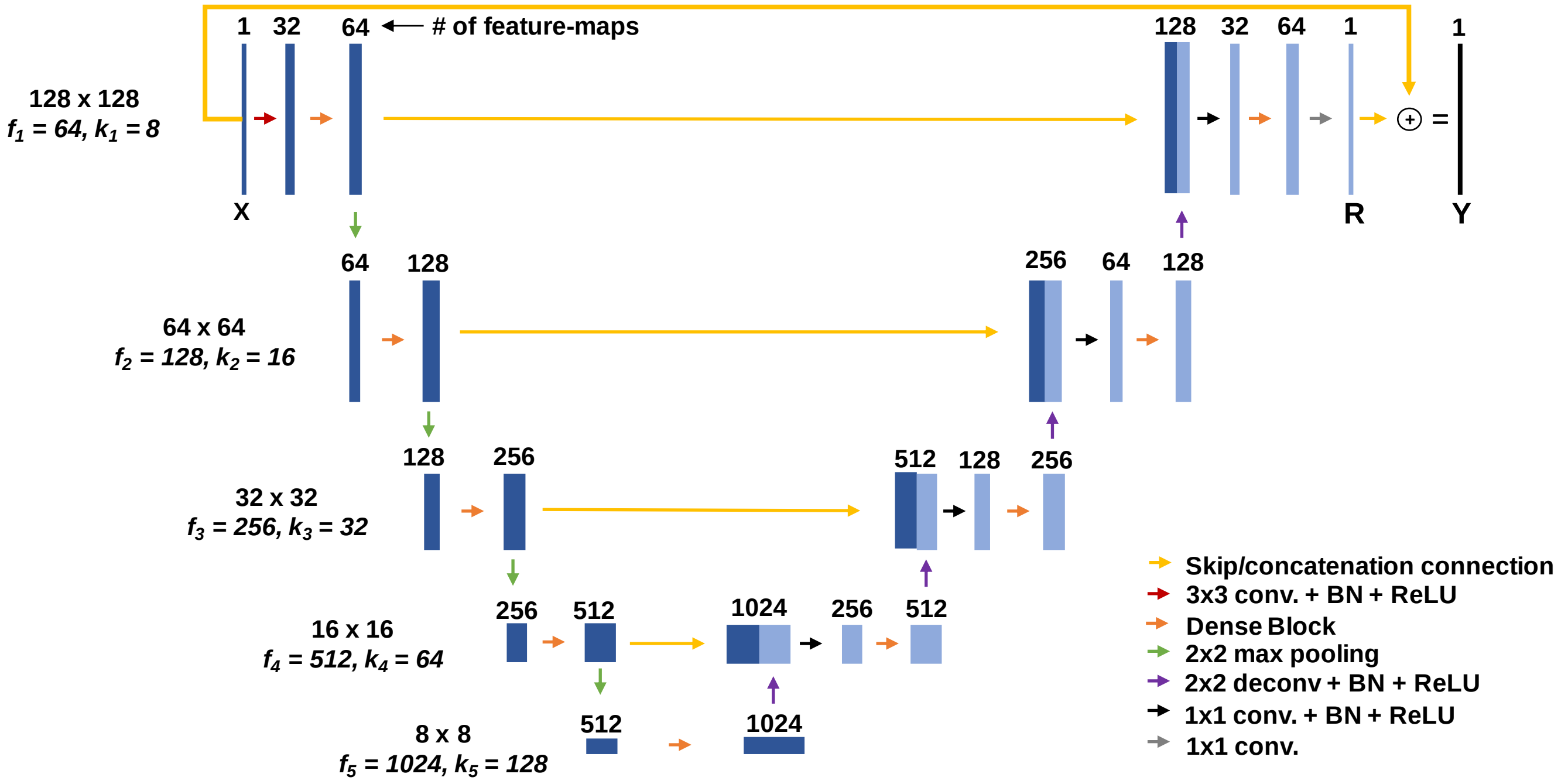

Fig. 3. Proposed FD-UNet architecture that incorporates dense connectivity [26] into the expanding and contracting path of the U-Net [19]. Hyperparameters for the illustrated architecture are $k_1 = 8$ and $f_1 = 64$ for an input image X of size 128x128 pixels.

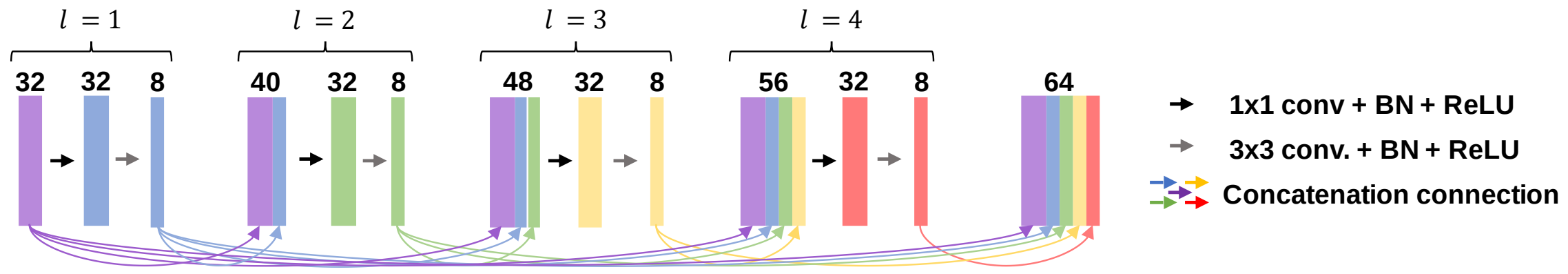

Fig. 4. Four layered dense block with $k_1 = 8$ and $F = 32$. Feature-maps from previous layers are concatenated together as the input to following layers.

For each spatial level, $s$, in the FD-UNet, a dense block with a growth rate, $k_s$, is used to learn a number of feature-maps, $f_s$. Initial values for $k_1$ and $f_1$ are hyperparameters defined by the user. $k_s$ is updated at each spatial level so that all dense blocks in the FD-UNet have the same number of convolutional layers to maintain computational efficiency. In our implementation, $k_s = 2^{s-1} \times k_1$ and $f_s = 2^{s-1} \times f_1$. Where the FD-UNet use dense blocks, the UNet have instead a sequence of two 3x3 convolution operations to learn feature-maps [27].

After each deconvolution operation, the upsampled feature-maps are concatenated channel-wise with feature-maps of similar size from the contracting path. These concatenation connections allow higher resolution features learned earlier in the network to be used in the upsampling process. However, this results in $2f_s$ feature-maps and cannot be reduced to the desired $f_s$ feature-maps using a dense block. To address this issue, the concatenated feature-maps are first reduced to $f_s/2$ feature-maps using a 1x1 convolution prior to each dense block in the expanding path.

In a dense block, earlier convolutional layers are connected to all subsequent layers via channel-wise concatenation [33], [34]. This means that the input to each layer in a dense block is the outputs from all previous layers concatenated together. Essentially, each layer learns additional feature-maps based on the "collective knowledge" gained by previous layers. This strategy increases the network's representational power through feature reuse. Features learned in earlier layers are passed forward and removes the need to learn redundant features and promotes learning a diverse set of features.

Furthermore, dense connectivity allows for deeper networks. For example, the FD-UNet has 82 convolution and deconvolution layers while the UNet has 23 layers. As the depth of the network increases, gradient information passes through many layers and can be lost before it reaches the earlier layers in a network leading to the vanishing gradient problem. Previous networks (e.g. ResNets and Highway Networks) addresses this problem by introducing short paths from earlier to later layers [40], [41]. Dense connectivity follows a similar principle but introduces many more connections to allow for

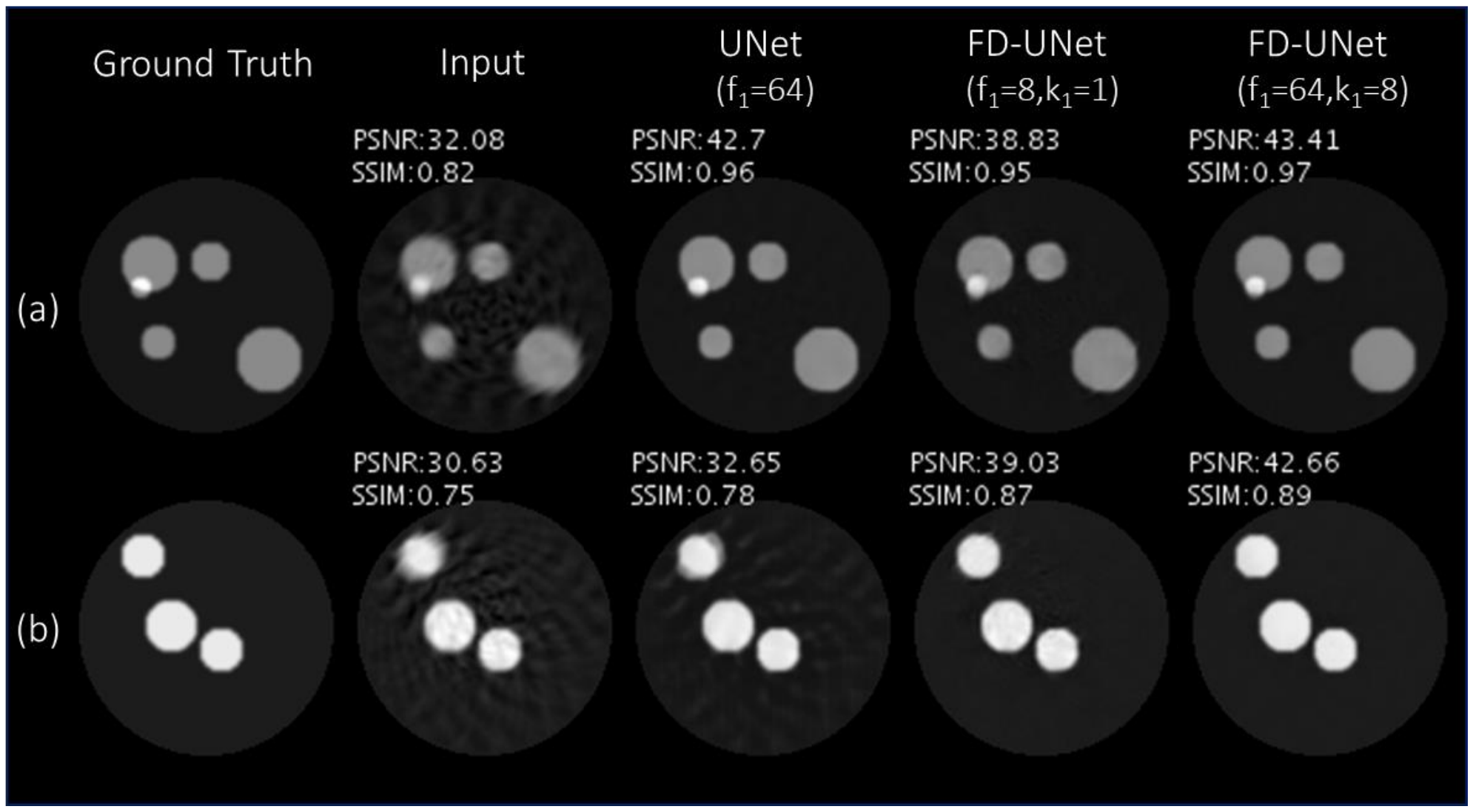

Fig. 5. Reconstructed circles images using TR, UNet, and FD-UNet with varying hyperparameters. (a) both CNNs recover a near artifact-free image. (b) example of the UNet reconstruction with residual background artifacts and the top-left circle has a distorted boundary.

TABLE I
AVERAGE PSNR AND SSIM FOR 2D CIRCLES DATASET (30 SENSORS)

| | $f_1 = 8$ $k_1 = 1$ | $f_1 = 16$ $k_1 = 2$ | $f_1 = 32$ $k_1 = 4$ | $f_1 = 64$ $k_1 = 8$ |
|---|---|---|---|---|
| TR | 32.48 ± 3.52 *0.75 ± 0.07* | | | |
| UNet | 33.77 ± 4.18<br>*0.78 ± 0.12*<br>487K<br>(0.94) | 34.48 ± 4.19<br>*0.79 ± 0.12*<br>1.9M<br>(1.05) | 34.70 ± 4.54<br>*0.79 ± 0.12*<br>7.8M<br>(1.55) | 34.84 ± 4.48<br>*0.79 ± 0.12*<br>31M<br>(2.94) |
| FD-UNet | 39.35 ± 3.19<br>*0.84 ± 0.08*<br>151K<br>(0.80) | 41.45 ± 3.28<br>*0.85 ± 0.07*<br>600K<br>(0.91) | 43.05 ± 3.27<br>*0.86 ± 0.07*<br>2.4M<br>(1.4) | 44.84 ± 3.42<br>*0.87 ± 0.07*<br>9.4M<br>(2.78) |

$f_1$ and $k_1$ are CNN hyperparameters. $k_1$ is only applicable to the FD-UNet. For each row, the following metrics are reported: PSNR, SSIM in italics, number of trainable parameters, and evaluation time in milliseconds for a single image in parenthesis.

gradient information to be efficiently backpropagated. This mitigates the vanishing gradient problem and allows for the network to be more easily trained.

As seen in Fig. 4., the $\ell^{th}$ layer in the dense block has an output with $k_s$ feature-maps and an input with $F + k_s \times (\ell - 1)$ feature-maps, where $F$ is the number of feature-maps of the initial input to the dense block. Features are learned through a sequence of a 1x1 and 3x3 convolution with batch normalization and rectified linear unit (ReLU) activation function [35], [36]. The 1x1 convolution is included to improve computational efficiency by reducing the input size to $F$ feature-maps prior to the more computationally expensive 3x3 convolution. Then $k_s$ features maps are learned from the reduced input using a 3x3 convolution. The final output of the dense block is the concatenation between the input and outputs from all dense block layers.

The proposed CNN architecture utilizes residual learning by adding a skip connection between the input and output [40], [42]. In residual learning, the CNN learns to map the input image $\boldsymbol{X}$ to a residual image $\boldsymbol{R} = \boldsymbol{Y} - \boldsymbol{X}$ and then recovers the target artifact-free image $\boldsymbol{Y}$ by adding the residual $\boldsymbol{R}$ to the input $\boldsymbol{X}$. Residual learning is shown to mitigate the vanishing gradient problem. The residual $\boldsymbol{R}$ often has a simpler structure than the original image and is easier for the CNN to learn [30].

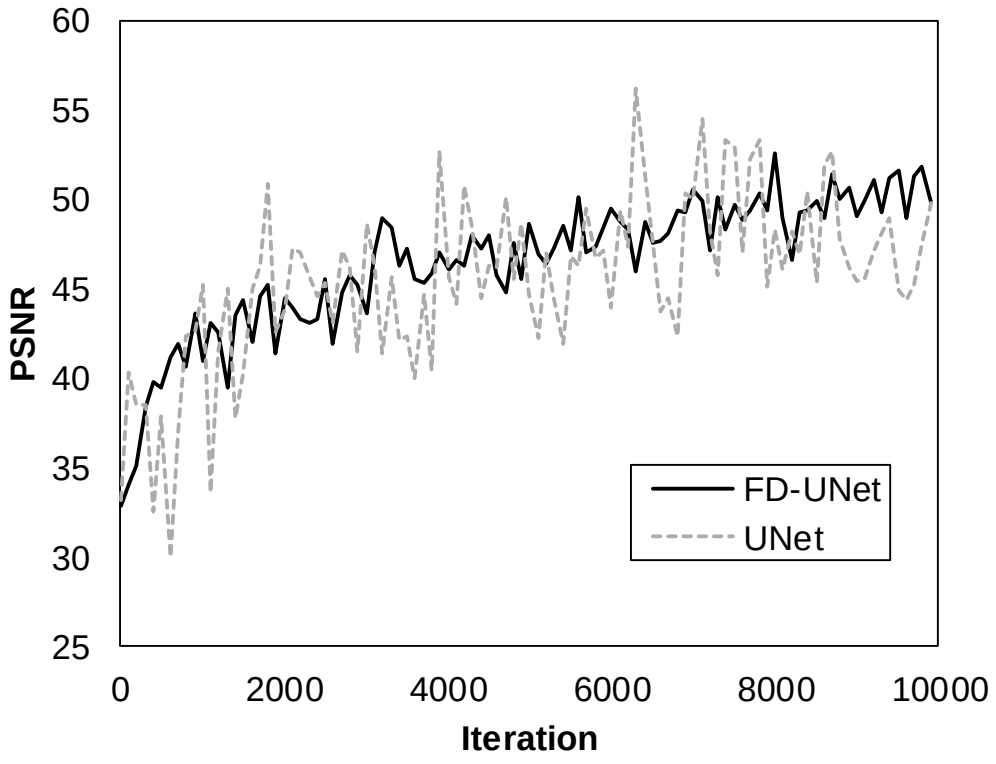

Fig. 6. Training loss in PSNR during the training phase for the FD-UNet ($f_1 = 64$, $k_1 = 8$) and UNet ($f_1 = 64$) on the circles training dataset (N=30 sensors).

## C. *Synthetic Data for Training and Testing*

Synthetic training and testing data is created using k-Wave, a MATLAB toolbox for simulating photoacoustic wave fields [43]. For each dataset generated, an initial photoacoustic source with a grid size of 128x128 pixels is defined. The medium is assumed to be non-absorbing and homogenous with a speed of sound of 1500 m/s. The sensor array has N detectors equally spaced on a circle with a radius of 60 pixels. Built-in functions of k-Wave are used to simulate sparse sampling of photoacoustic pressures. The TR method is then used to reconstruct an initial image containing artifacts from the sparsely sampled data.

Datasets are generated from three different synthetic phantoms (circles, Shepp-Logan, and vasculature) and an anatomically realistic vasculature phantom created from experimentally acquired micro-CT images of mouse brain vasculature. The phantoms are used to define an initial photoacoustic pressure source in k-Wave for creating simulated PAT images.

The circles dataset is comprised of simple phantoms that contain up to five circles with equal magnitude. The center location and radius for each circle are chosen randomly from a uniform distribution. This protocol is used to initially create a total of 1200 circles phantom images. We employed four-fold cross validation by dividing the images into four sets of a 1000 training images and 200 testing images. The images are used to initialize the photoacoustic pressure distribution to created simulated PAT image datasets for three levels of sampling sparsity (10, 15, and 30 detectors).

The Shepp-Logan and synthetic vasculature datasets are created using a data augmentation strategy. Training and testing images are procedurally generated from an original image with a size of 340x340 pixels for each phantom. Downsampled versions of these initial phantom images are shown as ground truth in Fig. 8. New images are created using the following steps. First, scaling and rotation is applied to the original image with a randomly chosen scaling factor (0.5 to 2) and rotation angle (0-359 degrees). Then a 128x128 pixels sub-image is randomly sampled from the transformed image. Finally, the sub-image is translated with a randomly selected vertical and horizontal shift (0-10 pixels) via zero-padding. Data augmentation allows for large sets of images with similar but different features to be easily created [44]. This strategy is used to generate a testing and fine-tuning dataset with 200 and 100 images, respectively, for each synthetic phantom. PAT images are then simulated using k-Wave with a sensor array of 30 detectors.

The anatomically realistic vasculature phantom is derived from a 3D volume of mouse brain vasculature that was experimentally acquired using micro-CT [45]. The original volume had a size of 260x336x438 pixels. The Frangi vesselness filter is applied to suppress background noise and enhance vessel-like features in the volume [46]. New images are created from the filtered volume following a similar data augmentation procedure as described for the synthetic phantoms. However, a 128x128x128 pixels sub-volume is instead randomly sampled from the transformed volume and is used to create a maximum intensity projection image by applying the max operator along the third dimension. Only a testing dataset with 200 images is generated from the mouse brain vasculature phantom. The corresponding training dataset with 1000 images is instead generated from the synthetic vasculature phantom. In order to create more complex synthetic images for training, the outputs from multiple iterations (up to five) of the data augmentation process are summed together. This enables the synthetic training images to have more a complex network structure with varying vessel sizes and orientation. PAT images of the synthetic and realistic vasculature phantoms are simulated at various levels of sampling sparsity (15, 30, and 45 detectors).

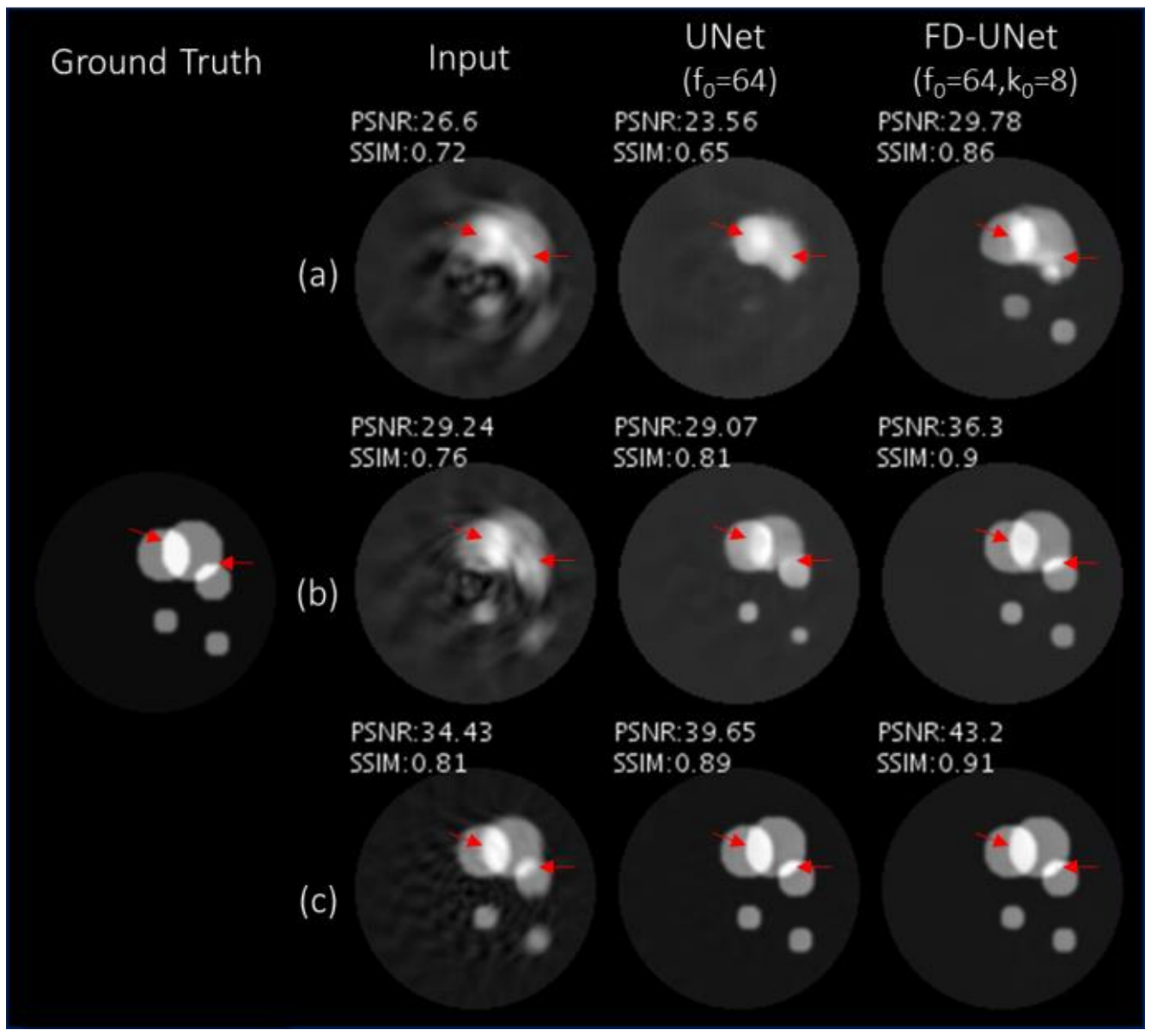

Fig. 7. Reconstructed circles images under different levels of sampling sparsity using (a) 10, (b) 15, and (c) 30 detectors. The red arrows point to a boundary that is blurred at more sparse sampling levels.

TABLE II
AVERAGE PSNR AND SSIM UNDER VARYING SAMPLING SPARSITY LEVELS

| # of Detectors | 10 | 15 | 30 |
|---|---|---|---|
| TR | 24.86 ± 3.18<br>*0.70 ± 0.05* | 27.30 ± 3.15<br>*0.72 ± 0.06* | 32.48 ± 3.52<br>*0.75 ± 0.07* |
| UNet | 24.69 ± 3.79<br>*0.72 ± 0.11* | 27.26 ± 3.94<br>*0.76 ± 0.11* | 34.84 ± 4.48<br>*0.79 ± 0.12* |
| FD-UNet | 32.59 ± 4.36<br>*0.83 ± 0.07* | 38.10 ± 4.20<br>*0.86 ± 0.07* | 44.84 ± 3.42<br>*0.87 ± 0.07* |

For each row, PSNR is shown as normal text on top while SSIM is shown as italicized text on the bottom. The CNN hyperparameters used are FD-UNet ($f_1 = 64$, $k_1 = 8$) and UNet ($f_1 = 64$)

### D. *Deep Learning Implementation*

The CNNs are implemented in Python 3.6 with TensorFlow v1.7, an open source library for deep learning [47]. Training and evaluation of the network is performed on a GTX 1080Ti NVIDIA GPU. The CNNs are trained for 10,000 iterations using a mean squared error loss function, learning rate of 1e-4, and a mini-batch size of three images.

## III. EXPERIMENTS AND RESULTS

The UNet and FD-UNet are compared over several experiments to determine if dense connectivity enables for more artifacts to be removed and hence an image with higher quality to be recovered. Image reconstruction quality is quantified using the peak signal-to-noise ratio (PSNR) and structural similarity index (SSIM) [48]. PSNR provides a global measurement of image quality whereas SSIM measures the similarity between local patterns of pixel intensities.

### A. *Circles Dataset*

In this initial experiment, the CNNs are both trained and tested using the circles dataset. This represents an ideal data scenario where the training and testing data are well-matched meaning the CNN had an opportunity to learn almost all of the features needed from the training data to perform well on the testing data. This ideal scenario provides a starting point for comparing the performance of the CNNs without limitations from data-related issues. Since the training and testing are derived from the same phantom in this experiment, four-fold cross validation is employed to increase confidence in the results observed.

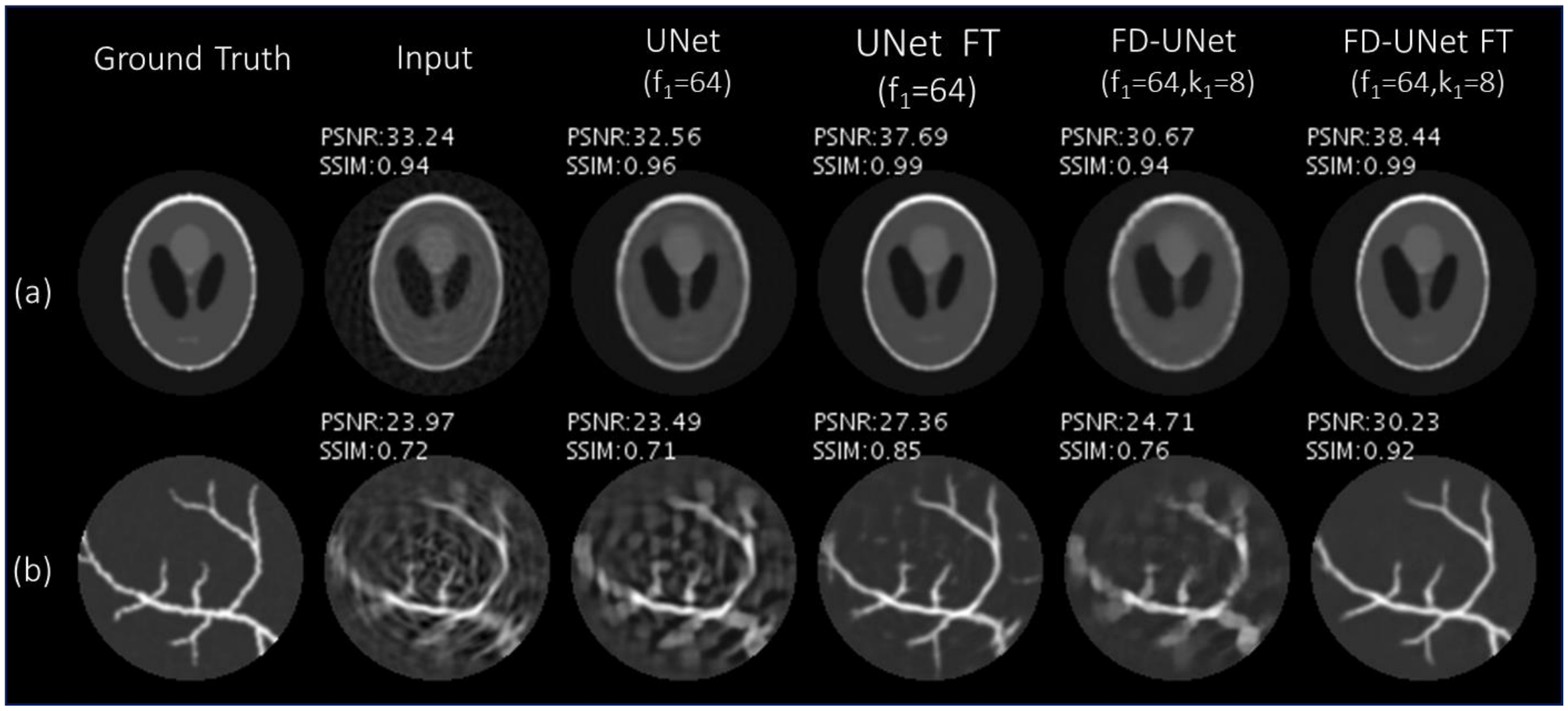

Fig. 8. Reconstructed images (30 sensors) of the (a) Shepp-Logan phantom and (b) vasculature phantom with and without fine-tuning (FT).

The CNNs' potential in learning to remove artifacts are evaluated by varying the hyperparameters $f_1$ (initial feature-maps learned) and $k_1$ (initial dense block growth rate). Increasing $f_1$ results in a wider CNN with more representational power and typically better performance. Results for the FD-UNet and UNet with varying model complexities for the circles dataset are shown in Table 1 and Fig. 5a. As expected, the initial TR reconstruction has severe artifacts and the lowest average PSNR and SSIM. Applying either CNN generally results in an improved and near artifact-free image. However, the FD-UNet outperforms and is more consistent in removing artifacts than the UNet. As seen in Fig. 5b., the FD-UNet removes majority of the artifacts but the UNet fails to remove artifacts on the boundary of the top-left circle and in the background. For all images in the testing dataset, there are no instances of the UNet outperforming the FD-UNet.

Dense connectivity improves model parameter efficiency and allows for a more compact CCN with better performance. As seen in Table 1, the FD-UNet requires fewer parameters (about a third) and has a higher average PSNR and SSIM compared to the UNet for each set of hyperparameters tested. The CNNs have similar average evaluation times with the FD-UNet being only slightly faster by a fraction of a millisecond. In the FD-UNet, a dense block is used in place of the two 3x3 convolutions in the UNet. While the dense block has eight different convolutional layers (four 1x1 and four 3x3), the input and output of each convolutional layer are relatively smaller. Thus, the convolutional layers in the dense block are computationally cheaper than those in the UNet resulting in the two CNNs having similar evaluation times.

TABLE III
AVERAGE PSNR AND SSIM FOR SHEPP-LOGAN AND VASCULATURE PHANTOM DATASET (30 DETECTORS)

| | Shepp-Logan | | Vasculature | |
|---|---|---|---|---|
| | Initial | Fine-tuned | Initial | Fine-tuned |
| TR | 32.50 ± 1.53<br>*0.87 ± 0.03* | | 24.79 ± 2.86<br>*0.66 ± 0.06* | |
| UNet | 31.69 ± 1.19<br>*0.93 ± 0.03* | 36.23 ± 2.46<br>*0.95 ± 0.04* | 24.40 ± 2.93<br>*0.66 ± 0.06* | 25.96 ± 2.85<br>*0.70 ± 0.11* |
| FD-UNet | 30.81 ± 0.97<br>*0.94 ± 0.01* | 38.24 ± 1.69<br>*0.97 ± 0.01* | 25.27 ± 2.16<br>*0.70 ± 0.05* | 31.30 ± 2.24<br>*0.82 ± 0.07* |

For each row, PSNR is shown as normal text on top while SSIM is shown as italicized text on the bottom. The CNN hyperparameters used are FD-UNet ($f_1$ = 64, $k_1$ = 8) and UNet ($f_1$ = 64)

Interestingly, the most compact FD-UNet ($f_1 = 8, k_1 = 1$) with fewer parameters and features learned outperforms the more complex UNet ($f_1 = 64$). This demonstrates that the FD-UNet, despite learning fewer features, is learning more relevant ones for artifact removal. In general, both CNNs have improved performance as $f_1$ and model complexity is increased. However, these improvements are diminishing because larger CNNs are more difficult to train and prone to overfitting. As seen in Fig. 6., the CNNs are trained for a total of 10,000 iterations but converge to a maximum by 8,000 iterations. The UNet loss appears to be more volatile compared to the FD-UNet loss.

The CNNs' ability to remove artifacts under varying levels of sampling sparsity are also evaluated. The goal of this experiment is to determine the extent of artifact severity that can be removed by each CNN. For each level of sampling sparsity, the CNNs are trained and tested on the corresponding datasets.

Results for the FD-UNet and UNet for different levels of sampling sparsity are shown in Table 2. As expected, decreasing the number of detectors used to sample the acoustic pressure results in more severe artifacts and a lower average PSNR and SSIM. The FD-UNet has a higher average PSNR and SSIM compared to the UNet for all levels of sampling sparsity tested. Reconstructed phantom images under different levels of sampling sparsity are shown in Fig. 7. When using 30 detectors, both CNNs perform well in removing artifacts from images reconstructed. At a sparser sampling level using 15 detectors, the FD-UNet recovers higher quality images than the UNet. For example, the boundaries of the circles as indicated by the red arrows in Fig. 7b. are blurred together in the UNet reconstruction but can be clearly distinguished in the FD-UNet reconstruction. Both CNNs are unable to reliably reconstruct the circles' boundaries at sparsity level using 10 detectors. Interestingly, the FD-UNet is able recover a reconstruction with a higher SSIM from a more corrupted initial image (10 detectors) than the UNet can from an initial image with less artifacts (30 detectors).

### B. *Shepp-Logan and Vasculature Phantom Dataset*

In the second experiment, the CNNs are initially trained on the circles dataset and tested on the Shepp-Logan and synthetic vasculature data. This represents a scenario in which the training and testing data are not necessarily well-matched. The circles and Shepp-Logan phantoms have many similar circular-like features and are fairly well-matched. However, the circles and synthetic vasculature phantom have significantly different features and are not well-matched. After initially training on the circles dataset, the CNNs are further trained for 5,000 iterations on either the Shepp-Logan or synthetic vasculature fine-tuning dataset. The purpose of this experiment is to evaluate the CNN's performance and ability to generalize when the training and testing datasets are not well-matched. Furthermore, the feasibility of training on a large poorly matched dataset and a smaller well-matched dataset is explored.

Results for the FD-UNet and UNet with and without fine-tuning for the Shepp-Logan and synthetic vasculature datasets are shown in Table 3 and Fig. 8. Both CNNs without fine-tuning have comparable performance and recover a high-quality albeit blurred reconstruction of the Shepp-Logan phantom as seen in Fig. 8a. However, they are not able to perform as well in the case of the of the synthetic vasculature phantom as seen in Fig. 8b. The general structure of the vessels can be clearly seen but appear to have circular-like features similar to the circles phantom training dataset. The FD-UNet does perform slightly better and removes more of the background artifacts.

As expected, fine-tuning with well-matched training data improves the CNNs' performance, especially in the case of the synthetic vasculature phantom. Both CNNs with fine-tuning recover a sharp and high-quality reconstruction of the Shepp-Logan phantom. Reconstructions of the synthetic vasculature no longer have the circle-like appearance. While both CNNs improve the initial TR reconstruction, the FD-UNet is able to remove more artifacts and outperform the UNet as evidenced by its higher average PSNR and SSIM for both synthetic phantoms.

### C. *Mouse Brain Vasculature Dataset*

In the third experiment, the CNNs are trained on the more complex synthetic vasculature phantom dataset and tested on the mouse brain vasculature dataset. In this scenario, the datasets are fairly well-matched, but there are likely features in the anatomically realistic brain vasculature dataset that are not present in the synthetic vasculature dataset. The purpose of this experiment is to evaluate the feasibility of training the CNNs on synthetic phantom images for removing artifacts from anatomically realistic vasculature images under multiple levels of sampling sparsity.

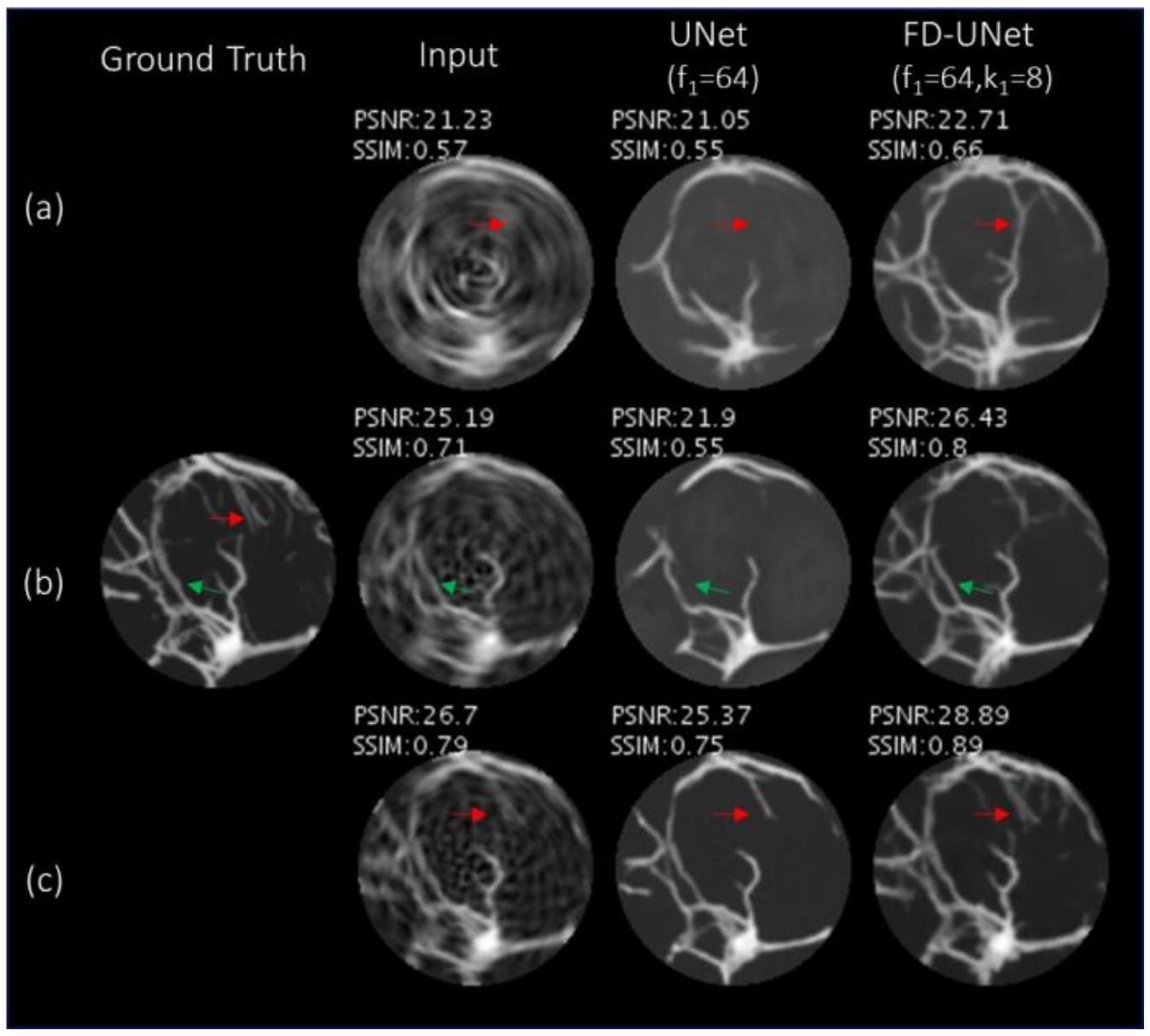

Fig. 9. Examples of reconstructed mouse brain vasculature images for sampling sparsity levels with (a) 15, (b) 30, and (c) 45 detectors. Red and green arrows point to features present in the FD-UNet but missing in the UNet reconstruction.

TABLE IV
AVERAGE PSNR AND SSIM UNDER VARYING SAMPLING SPARSITY LEVELS FOR MOUSE BRAIN VASCULATURE DATASET

| # of Detectors | 15 | 30 | 45 |
|---|---|---|---|
| TR | 19.77 ± 0.96<br>*0.58 ± 0.05* | 22.89 ± 1.13<br>*0.70 ± 0.05* | 25.56 ± 1.28<br>*0.78 ± 0.05* |
| UNet | 20.21 ± 1.19<br>*0.60 ± 0.07* | 22.15 ± 2.35<br>*0.68 ± 0.11* | 25.07 ± 2.09<br>*0.76 ± 0.11* |
| FD-UNet | 21.12 ± 1.18<br>*0.65 ± 0.04* | 25.13 ± 1.36<br>*0.82 ± 0.03* | 28.47 ± 1.39<br>*0.89 ± 0.03* |

For each row, PSNR is shown as normal text on top while SSIM is shown as italicized text on the bottom. The CNN hyperparameters used are FD-UNet ($f_1 = 64, k_1 = 8$) and UNet ($f_1 = 64$)

As seen in Table 4, there are no significant quantitative changes in PSNR and SSIM between the UNet and TR reconstructions for all levels of sampling sparsity tested. However, the UNet does remove majority of the background artifacts and qualitatively appears better than the TR reconstruction as shown in Fig. 9. No quantitative improvement is observed because the UNet only recovers larger vessels and is missing many of the smaller features. The FD-UNet outperforms the UNet and improves the average PSNR and SSIM. It recovers many of the smaller details that are missing in the UNet reconstruction as shown by the green arrows in Fig. 9b. The performance of the CNNs is heavily dependent on the image quality of the TR reconstruction. Features that are missing in the initial reconstruction are also typically missing or incorrectly reconstructed by the CCNs as shown by the red arrows in Fig. 9a.

## IV. Discussion

In this work, we propose a modified CNN architecture for removing artifacts from 2D PAT images reconstructed from sparse data. Results from the experiments performed consistently show that the FD-UNet is superior to the standard UNet for artifact removal and image enhancement. Dense connectivity strongly encourages feature reuse and improves information flow throughout the network. The benefits in using this connectivity pattern can be observed in Fig. 5. The most compact FD-UNet ($f_1 = 8$) outperforms the more complex UNet ($f_1 = 64$) despite learning fewer features and requiring only a fraction of the parameters. This demonstrates that the FD-UNet is learning more relevant features for artifact removal, and the ability to reuse those features throughout the network greatly improves the CNN's performance. Furthermore, dense connectivity has a regularizing effect that reduces the likelihood of overfitting to the training data. As seen in Fig. 6., both CNNs converge to a similar PSNR during training yet the FD-UNet outperforms the UNet in testing data. This is likely due to the UNet overfitting to the training data and failing to lean features that generalize well. Furthermore, the UNet training loss is more volatile relative to that of the FD-UNet indicating that the UNet is overfitting to previously observed training examples.

A limitation in using deep learning for artifact removal is that the CNN requires a large training dataset to learn the appropriate weights and features needed to perform well. This limitation can be addressed using computational models (e.g. k-Wave) and synthetic phantoms to generate arbitrarily large datasets for training. However, there remains a challenge in generating a training dataset with all the image features likely to be observed in the testing dataset. This requirement for well-matched training and testing data is demonstrated in the second experiment. As seen in Fig. 8, the CNNs having trained only on images of circles can recover good reconstructions of the Shepp-Logan phantom but not of the synthetic vasculature phantom. Their performance is improved after fine-tuning with a small dataset of synthetic vasculature images. These results provide evidence that it is feasible to initially train the CNN using a poorly matched dataset and then fine-tuned using a small well-matched dataset. This strategy may be useful when only a few relevant experimental training images are available.

In the third experiment, the FD-UNet is trained on the synthetic vasculature dataset and tested on the mouse brain vasculature dataset. While both CNNs remove majority of the background artifacts and reliably recover the larger vessels, the FD-UNet typically recovers more of the smaller vessels than the UNet as seen in Fig. 9. As fewer detectors are used for sampling, the artifacts become increasingly severe in the TR reconstruction and image quality is degraded. A limitation in the post-processing approach is that the CNN's performance strongly depends on the quality of the TR reconstruction. Image features severely obscured by artifacts or missing in the TR reconstruction are likely to be reconstructed incorrectly or missing in the CNN reconstruction. Information is lost as a result of sparse sampling, but the initial step of reconstructing an image from sensor data also discards potentially useful information and introduces artifacts. It may be possible to recover some of the smaller vessels if the CNN is used to directly reconstruct the sensor data into an image.

## V. Conclusion

In this paper, we propose a modified CNN architecture termed FD-UNet for removing artifacts from 2D PAT images reconstructed from sparse data. We compare the FD-UNet and the UNet using datasets generated from synthetic phantoms (circles, Shepp-Logan, and vasculature) and an anatomically realistic mouse brain vasculature dataset. The FD-UNet is demonstrated to be superior and more compact CNN for removing artifacts and improving image quality.

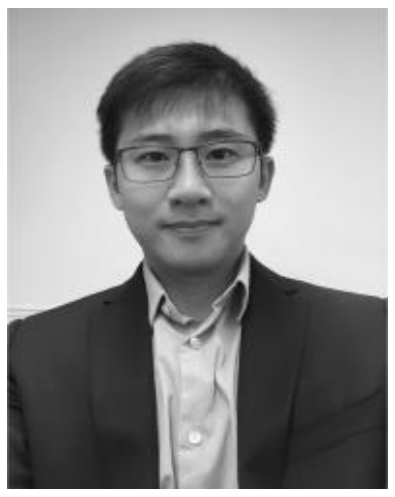

**Steven Guan** is currently a Ph.D. candidate at George Mason University in the Bioengineering Department. He received a B.S. in chemical engineering, B.A. in physics, and M.S. in biomedical engineering from the University of Virginia.  He is working as a senior data scientist for the MITRE corporation and supports multiple government agencies. His current areas of research interest include applying deep learning techniques for medical imaging classification, segmentation, and reconstruction.

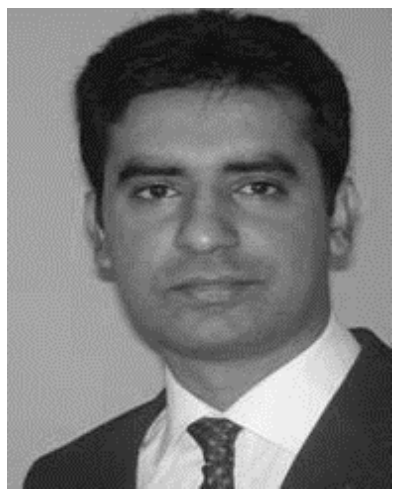

**Amir A. Khan** received his Ph.D. in electrical engineering from Grenoble Institute of Technology, Grenoble INP, France in 2009. He is currently with the Department of Bioengineering, George Mason University, VA, USA. He has previously worked with EDF, France as measurement and signal processing engineer and National University of Sciences and Technology, Pakistan as Assistant Professor. His research interests include data analytics and machine learning for biomedical and industrial applications. His current research focuses on analyzing multi-modal imaging data to develop clinical bioinformatics and machine learning tools for assessing the relationship between carotid artery disease, cerebral perfusion and cognitive impairment. He is also interested in developing novel deep-learning techniques for medical image segmentation, enhancement and reconstruction.

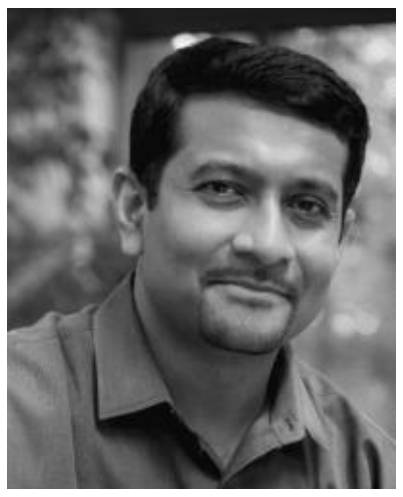

**Dr. Siddhartha Sikdar** is currently a Professor in the Department of Bioengineering at George Mason University. He is the Director of the Center for Adaptive Systems of Brain-Body Interactions (CASBBI). Dr. Sikdar’s research group within CASBBI conducts translational research using imaging to investigate brain-body interactions in a number of clinical conditions of major public health significance, such as chronic pain, stroke, spinal cord injury, and amputation. Dr. Sikdar obtained his PhD in Electrical Engineering from University of Washington, Seattle in 2005. He received a postdoctoral fellowship from the American Heart Association. Dr. Sikdar has been a recipient of the NSF CAREER Award, the Volgenau School of Engineering Rising Star Award, and Mason’s Emerging Researcher/Scholar/Creator Award.

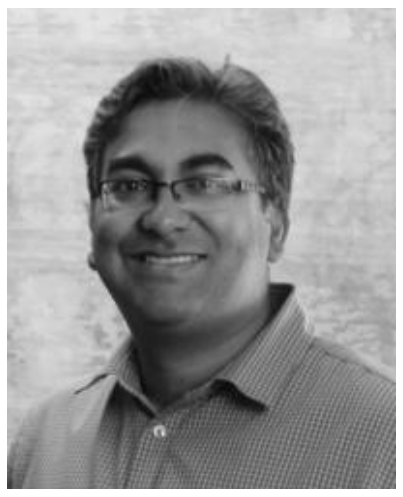

**Parag V. Chitnis** (M’08) has been an Assistant Professor in the Department of Bioengineering at George Mason University since 2014. He received a B.S. degree in engineering physics and mathematics from the West Virginia Wesleyan College, Buckhannon, WV, in 2000. He received M.S. and Ph.D. degrees in mechanical engineering from Boston University in 2002 and 2006, respectively. His dissertation focused on experimental studies of acoustic shock waves for therapeutic applications. After a two-year postdoctoral fellowship at Boston University involving a study of bubble dynamics, Dr. Chitnis joined Riverside Research as a Staff Scientist in 2008, where he pursued research in high-frequency ultrasound imaging, targeted drug delivery, and photoacoustic imaging. His current areas of research interest include therapeutic ultrasound and neuromodulation, photoacoustic neuro-imaging, and deep-learning strategies for photoacoustic tomography. He currently serves as an Associate Editor for Ultrasonic Imaging and a reviewer for NIH and NSF grant panels.